\documentclass[conference]{IEEEtran}
\IEEEoverridecommandlockouts
\usepackage{cite}
\usepackage{amsmath,amssymb,amsfonts}
\usepackage{algorithmic}
\usepackage{graphicx, multirow}
\usepackage{textcomp}
\usepackage{xcolor}
\usepackage[caption = false]{subfig}
\usepackage{float}
\newcommand*{\tran}{^{\mkern-1.5mu\mathsf{T}}}

\def\BibTeX{{\rm B\kern-.05em{\sc i\kern-.025em b}\kern-.08em
    T\kern-.1667em\lower.7ex\hbox{E}\kern-.125emX}}
\usepackage{etoolbox}
\makeatletter
\patchcmd{\@makecaption}
{\scshape}
{}
{}
{}
\makeatother
\def\tablename{Table}
\begin{document}

\title{
Multimodal Image Super-resolution via \\Deep Unfolding with Side Information 
}

\author{\IEEEauthorblockN{Iman Marivani, Evaggelia Tsiligianni, Bruno Cornelis, Nikos Deligiannis}
	\IEEEauthorblockA{\textit{Department of Electronics and Informatics, Vrije Universiteit Brussel, Pleinlaan 2, B-1050 Brussels, Belgium} \\
		\textit{imec, Kapeldreef 75, B-3001 Leuven, Belgium}\\}
	
}

\maketitle

\begin{abstract}
Deep learning methods have been successfully applied to various computer vision tasks.
However, existing neural network architectures do not per se incorporate domain knowledge about the addressed problem,
thus, understanding what the model has learned is an open research topic. 
In this paper, we rely on the unfolding of an iterative algorithm for sparse approximation with side information, 
and design a deep learning architecture for multimodal image super-resolution
that incorporates sparse priors and effectively utilizes information from another image modality.
We develop two deep models performing reconstruction of a high-resolution image of a target image modality 
from its low-resolution variant with the aid of a high-resolution image from a second modality.  
We apply the proposed models to super-resolve near-infrared images 
using as side information high-resolution RGB\ images. 
Experimental results demonstrate the superior performance of the proposed models 
against state-of-the-art methods including unimodal and multimodal approaches. 
\end{abstract}

\begin{IEEEkeywords}
Image super-resolution, sparse coding, multimodal deep learning, designing neural networks.
\end{IEEEkeywords}
\vspace{-0.1cm}

\section{Introduction}
Image super-resolution (SR) refers to the recovery of a high-resolution (HR) image from its low-resolution (LR) version. 
The problem is severely ill-posed 
and a common approach for its solution considers the use of sparse priors~\cite{sparse1,sparse2,sparse3}. 
For example, the method presented in~\cite{sparse1}
is based on the assumption that the LR and HR images have joint sparse representations 
with respect to some dictionaries. 
Nevertheless, sparsity based methods result in complex optimization problems, 
which is a significant drawback in large-scale settings.

Accounting for the high computational cost of numerical optimization algorithms, 
deep neural networks have been successfully applied to image SR 
achieving state-of-the-art performance~\cite{10prim,11prim,15prim,17prim,19prim}. 
Deep learning methods rely on large datasets to learn a non-linear transformation
between the LR and HR image spaces. 
The methods are efficient at inference by shifting the computational load to the training phase. 
However, most of the existing deep models do not integrate domain knowledge about the problem
and cannot provide theoretical justifications for their effectiveness.
A different approach was followed in the recent work of~\cite{Huang},  
which relies on a deep unfolding architecture referred to as LISTA~\cite{LISTA}.
LISTA introduced the idea of translating an iterative numerical algorithm for sparse approximation 
into a feed-forward neural network.
By integrating LISTA into their network architecture, 
the authors of~\cite{Huang} managed to incorporate sparse priors into the deep learning solution.

In many image processing and machine vision applications 
a reference HR image from a second modality is often available~\cite{MSR2,DJF}.
The recovery of an HR image from its LR variant
with the aid of another HR image from a different image modality
is referred to as multimodal image SR~\cite{DJF,MSR}.
Several studies have investigated sparse representation models 
as well as deep learning methods for multimodal image SR~\cite{MSR, DJF, MSR1, MSR2, MSR3, MSR4}.

In this paper, we propose a deep network architecture that incorporates sparse priors
and effectively utilizes information from another image modality to perform multimodal image SR.
Inspired by~\cite{Huang}, 
the proposed deep learning model relies on a deep unfolding method 
for sparse approximation with side information.
Our contributions are threefold:
\begin{enumerate}
\item We address multimodal image SR as a problem of sparse approximation with side information  
and formulate an appropriate $\ell_1$-$\ell_1$ minimization problem for its solution.
\item We design a core neural network model for patch-based image SR, employing 
a recently proposed deep sparse approximation model~\cite{EVA}, referred to as LeSITA, 
to integrate information from another HR image modality in the solution.
\item We propose two novel deep neural network architectures 
for multimodal image SR that employ the LeSITA-based SR model,  
achieving superior performance against state-of-the-art methods.
\end{enumerate}
The proposed models are used to super-resolve near-infrared (NIR) LR images given RGB HR images. 
The performance of the proposed models is demonstrated by experimental results.

The rest of the paper is organized as follows: 
In Section~\ref{sec:back} we present the necessary background and related work.
Section~\ref{sec:proposed} explains the details of the proposed model architectures
for multimodal image SR. 
Section~\ref{sec:experiment} presents the experimental results, 
and Section~\ref{sec:conclusion} concludes the paper.

%%%%%%%%%%%%%%%%%%%%%%%%%%
\section{Background and Related Work}
\label{sec:back}
%%%%%%%%%%%%%%%%%%%%%%%%%%
%--------------------------------------------------------------------%
\subsection{Single Image SR with Sparse Priors}
\label{sec:sparseback}
%--------------------------------------------------------------------%
Let $Y \in \mathbb{R}^{n_1^{(y)} \times n_2^{(y)}}$ be an LR image 
obtained from an HR image $X \in \mathbb{R}^{n_1^{(x)} \times n_2^{(x)}}$.
According to~\cite{sparse1},
the transformation of an HR image to an LR image can be modeled
as a blurring and downscaling process expressed by $Y = BAX$, 
where $A$ and $B$ denote blurring and downscaling operators, respectively. 
Under this assumption, an $n_y$-dimensional (vectorized) patch $y$ from the bicubic-upscaled LR image
exhibits a common sparse representation with the corresponding patch $x$ from the HR image 
w.r.t. different over-complete dictionaries $D_y \in \mathbb{R}^{n_y \times n_{\alpha}}$, $D_x \in \mathbb{R}^{n_x \times n_{\alpha}}$, 
that is, $y = D_y\alpha$ and $x = D_x\alpha$,
where $\alpha \in \mathbb{R}^{n_{\alpha}}$.
$D_y$, $D_x$ can be jointly learned with coupled dictionary learning techniques~\cite{sparse1}.
Therefore, the problem of computing the HR patch $x$ given the LR patch $y$ can be formulated as
\begin{equation}
\label{eq:sideeq}
x = D_x\alpha, \quad \text{s.t.} \quad \alpha = \arg \min_{v\in \mathbb{R}^{n_{\alpha}}} \| y - D_yv \|_2^2 + \lambda\|v\|_1,
\end{equation}
where $\lambda$ is a regularization parameter, 
and $\|v\|_1 = \sum_{i=1}^{n_{\alpha}}|v_i|$ is the $\ell_1$-norm, which promotes sparsity. 
Several numerical optimization methods have been proposed 
for the solution of~\eqref{eq:sideeq}~\cite{tropp2010computational}.

\begin{figure}[t!]
	\centering
	\includegraphics[width=0.33\textwidth]{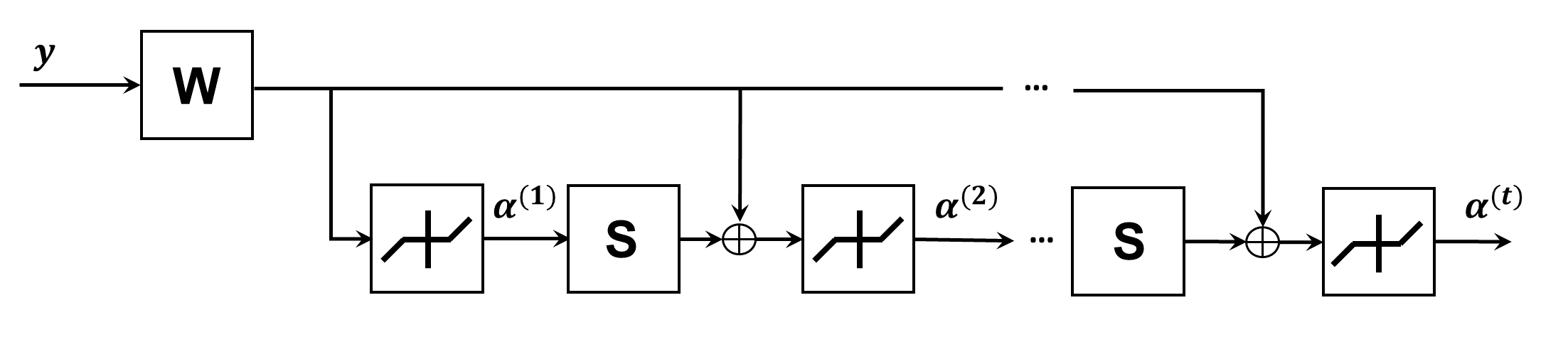}\\
	\caption{Deep learning for sparse coding with LISTA~\cite{LISTA}. }
	\label{fig:LISTA}
\end{figure}

In order to account for the high computational cost of numerical algorithms, 
the seminal work in~\cite{LISTA} translated a proximal algorithm, namely ISTA~\cite{ISTA}, 
into a neural network form referred to as LISTA.
Each layer of LISTA implements an iteration of ISTA according to 
\begin{equation}
\label{eq:lista}
\alpha^{t+1} = \phi_{\gamma}(S\alpha^t + Wy),
\end{equation}
where $W$, $S$, and $\gamma$ are learnable parameters, and
$\phi_{\gamma}$ is the soft thresholding operator~\cite{ISTA} 
expressed by the component-wise shrinkage function
$\phi_{\gamma}(u_i) = \text{sign}(u_i)\max\{|u_i| - \gamma, 0\}$,  $i=1, \dots , n_{\alpha}$,
which acts as a nonlinear activation function. $W$ and $S$ are initialized as: $S=I- \frac{1}{L} D\tran D$, $W=\frac{1}{L} D\tran $.
The network is depicted in Fig.~\ref{fig:LISTA}.

By employing the model presented in~\cite{sparse1} 
and leveraging the fast computations of sparse codes performed by LISTA,
the authors of~\cite{Huang}  proposed a deep neural network model for single-modal image super resolution 
that incorporates sparse priors.

%--------------------------------------------------------------------%
\subsection{Sparse Coding with Side Information via Deep Learning}
\label{sec:sparseback}
%--------------------------------------------------------------------%
The deep unfolding idea presented in~\cite{LISTA}, also explored in~\cite{xin2016maximal, borgerding2017amp, prox},
introduced a new methodology for the design of neural networks,
enabling the network structure to incorporate domain knowledge about the addressed problem.
Following similar principles, 
we recently proposed a deep learning architecture to solve the  
sparse approximation problem with the aid of side information~\cite{EVA}.
Our approach relies on a proximal algorithm for  $\ell_1$-$\ell_1$ minimization.
Specifically, suppose that we want to find a sparse approximation $\alpha \in \mathbb{R}^{n_{\alpha}}$ of a signal $y \in \mathbb{R}^{n_y}$
w.r.t. an over-complete dictionary $D_y \in \mathbb{R}^{n_y \times n_{\alpha}}$,
given a side information signal $\tilde{\alpha} \in \mathbb{R}^{n_{\alpha}}$ that is correlated to $\alpha$.
Then, by solving the $\ell_1$-$\ell_1$ minimization problem
\begin{equation}
\label{l1l1}
{\alpha}  = \arg \min_{v\in \mathbb{R}^{n_{\alpha}}}  \frac{1}{2}\| y - D_y v\|^2_2+\lambda\big(\|v\|_1 +\|v-\tilde{\alpha}\|_1\big),
\end{equation}
we obtain a solution that is of higher accuracy compared to~\eqref{eq:sideeq},
as long as certain conditions concerning the similarity between $\alpha$ and $\tilde{\alpha}$ hold~\cite{NikosIT, nikosGlobsip}.

A proximal algorithm that solves~\eqref{l1l1} performs the following iterations~\cite{EVA ,Bach2012}:
\begin{equation}
\label{eq:lesita}
\alpha^{t+1} = \xi_{\mu}\big(\alpha^t- \frac{1}{L}D\tran(D\alpha^t -y)\big),
\end{equation}
where $L$ is the Lipschitz constant of $\nabla \| y - D_y v\|^2_2$ and $\mu=\frac{\lambda}{L}$.
$\xi_\mu(u;\tilde{\alpha})$ is the proximal operator expressed by:
\begin{enumerate}
	\item{for $\tilde{\alpha}_i \ge 0$, $i=1, \dots, n_{\alpha}$:
		\begin{equation}
		\label{eq:prox1}
		\xi_{\mu}(u_i;\tilde{\alpha}_i) =  
		\begin{cases}
		u_i + 2\mu, \, \; \; \, \; \; \qquad  u_i < -2\mu  \\
		0, \quad \qquad \; \; \, \; \;  -2\mu \le u_i \le 0\\
		u_i , \qquad \qquad \, \; \; \;  0 < u_i < \tilde{\alpha}_i \\  
		\tilde{\alpha}_i , \qquad \;  \, \tilde{\alpha}_i \le u_i \le \tilde{\alpha}_i +2\mu \\
		u_i - 2\mu , \, \qquad  u_i \ge \tilde{\alpha}_i +2\mu
		\end{cases}
		\end{equation}
	}
	\item{for $\tilde{\alpha}_i < 0$, $i=1, \dots, n_{\alpha}$:
		\begin{equation}
		\label{eq:prox2}
		\xi_{\mu}(u_i;\tilde{\alpha}_i) =  
		\begin{cases}
		u_i + 2\mu,  \qquad  u_i < \tilde{\alpha}_i -2\mu  \\
		\tilde{\alpha}_i , \qquad \;   \tilde{\alpha}_i - 2\mu \le u_i \le \tilde{\alpha}_i \\
		u_i , \qquad \; \; \; \qquad  \tilde{\alpha}_i < u_i < 0 \\ 
		0, \qquad \quad \qquad  0 \le u_i \le 2\mu \\
		u_i - 2\mu , \, \qquad \qquad  u_i \ge 2\mu  
		\end{cases}.
		\end{equation}
	}
\end{enumerate}

By setting $Q=I- \frac{1}{L} D\tran D$, $R=\frac{1}{L} D\tran $,  \eqref{eq:lesita} takes the form:
\begin{equation}
\label{eq:lesitaaa}
\alpha^{t+1} = \xi_\mu(Q\alpha^t + Ry;\tilde{\alpha}).
\end{equation}
A feed forward neural network performing operations according to \eqref{eq:lesitaaa}
can learn sparse codes with the aid of side information.
$Q$, $R$, and  $\mu$ are parameters learned from data. 
Compared to LISTA~\cite{LISTA}, 
the network in~\cite{EVA}---which we call Learned Side-Information-driven iterative soft Thresholding Algorithm (LeSITA)---incorporates a new activation function~$\xi_\mu(\cdot;\tilde{\alpha})$ integrating the side information signal $\tilde{\alpha}$ into the sparse representation learning process.
\begin{figure}[t!]
	\centering
	\includegraphics[width=0.35\textwidth]{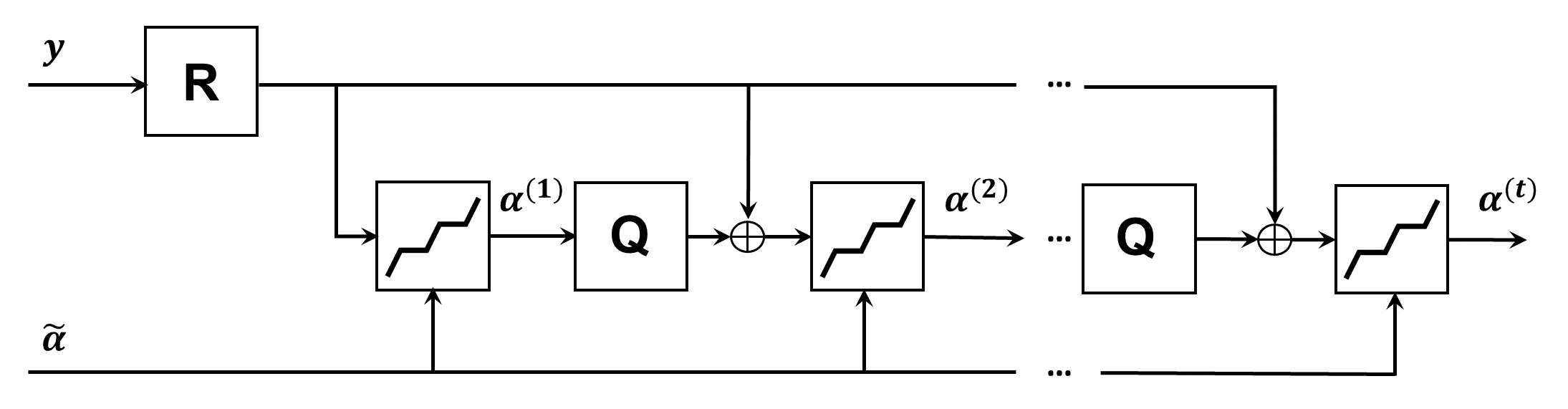}\\
	\caption{Deep learning for sparse coding with side information using LeSITA~\cite{EVA}. }\label{fig:LeSITA}
\end{figure}

%%%%%%%%%%%%%%%%%%%%%%%%%%
\section{The Proposed Method}
\label{sec:proposed}
%%%%%%%%%%%%%%%%%%%%%%%%%%
\subsection{Multimodal Image SR with Sparse Priors}
The problem of multimodal image super-resolution  
concerns the reconstruction of an HR image $X \in \mathbb{R}^{n_1^{(x)} \times n_2^{(x)}}$ 
from an LR image $Y \in \mathbb{R}^{n_1^{(y)} \times n_2^{(y)}}$,
given an HR reference or guidance image $Z \in \mathbb{R}^{n_1^{(z)} \times n_2^{(z)}}$ from another modality. 
In this work, we utilize the reference image $Z$ as side information
and leverage our previous work~\cite{EVA} to build a deep network performing multimodal image SR. 

We follow the sparsity-based model presented in~\cite{sparse1} 
and assume that a  (vectorized) patch $y \in \mathbb{R}^{n_y}$ from the bicubic-upscaled LR  image $Y$
and an HR patch $x \in \mathbb{R}^{n_x}$ from the high resolution image $X$  
share the same sparse representation $\alpha \in \mathbb{R}^{n_{\alpha}}$ 
under over-complete dictionaries $D_y \in \mathbb{R}^{n_y \times n_{\alpha}}$ and $D_x \in \mathbb{R}^{n_x \times n_{\alpha}}$, respectively. 
If the images of the target and the guidance modalities are highly correlated,
we can also assume that the reference patch $z \in \mathbb{R}^{n_z}$
has a sparse representation $\tilde{\alpha} \in \mathbb{R}^{n_{\alpha}}$ 
under a dictionary $D_z \in \mathbb{R}^{n_z \times n_{\alpha}}$,
which is similar to $\alpha$, for example, by means of the $\ell_1$ norm. 
Then, the multimodal image super-resolution problem
can be formulated  as an $\ell_1$-$\ell_1$ minimization problem of the form~\eqref{l1l1}.

By employing LeSITA to solve~\eqref{l1l1},
we can design an end-to-end multimodal deep learning architecture 
to perform super-resolution of the input LR image $Y$ with the aid of a reference HR image $Z$.
The network architecture incorporates sparse priors and exploits the correlation between the two available modalities. 
The proposed framework is presented next.
\begin{figure}[t!]
	\centering
	\includegraphics[width=0.33\textwidth]{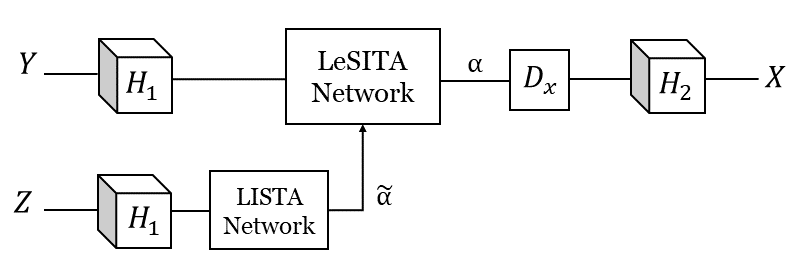}\\
	\caption{The proposed DMSC model for multimodal image SR consists of (i) a LeSITA encoder 
	computing a latent representation of the LR/HR images of the main modality,
	using side information from the guidance modality provided by (ii)  a LISTA encoder. 
	A linear decoder recovers the HR patch from the latent representation.
	The convolutional layers $H_1$,  $H_2$ perform patch extraction 
	and patch aggregation operations, respectively.} 
	\label{fig:base}
\end{figure}
%--------------------------------------------------------------------%
\subsection{DMSC: Deep Multimodal Sparse Coding Network} 
\label{sec:base}
%--------------------------------------------------------------------%
LeSITA can learn sparse codes of a target image modality
using side information from another correlated image modality.
The side information needs to be a sparse signal similar to the target sparse code.
To obtain sparse codes of the guidance modality,
our architecture also includes a LISTA subnetwork. 
The proposed core model consists of the following three components: 
(i) a LeSITA encoder that computes a sparse representation of an LR image patch of the target modality
using side information,
(ii) a LISTA subnetwork that produces a sparse representation of the available HR patch 
from the guidance image, and
(iii) a linear decoder that reconstructs the HR image patch of the main modality using the 
sparse representation obtained from LeSITA.

LeSITA  computes a sparse representation $\alpha$ 
of the LR patch $y$ according to \eqref{eq:lesitaaa}.
LISTA accepts as input the reference patch $z $ 
and performs a nonlinear transformation according to $\tilde{\alpha}^{t+1} = \phi_{\gamma}(S\tilde{\alpha}^t + Wz)$
to produce a side information signal for LeSITA.
Given the sparse representation $\alpha$ produced by LeSITA, 
the HR patch $x$ can be recovered by a linear decoder according to $x = D_x \alpha$.
$D_x$ is a learnable dictionary.
By training the network end-to-end, an LR/HR transformation that relies on 
the joint representations provided by LeSITA can be learned.

The proposed core model successfully performs multimodal image SR at a patch level.
Nevertheless, our goal is to design a network 
that accepts as input the entire images $Y$ and $Z$ 
and outputs the HR image $X$.    
To this end, we add three more layers to the network as follows. 
A convolutional layer $H_1$ consisting of $m$ filters of size $k \times k$ is added before the LeSITA encoder 
to extract $m$-dimensional feature vectors from the LR image corresponding to patches of size $k \times k$. 
A similar layer is added before the LISTA branch 
to ensure that the side information patches stay aligned with the LR patches. 
Finally, a convolutional layer $H_2$ is added after the decoder $D_x$ 
to aggregate the reconstructed HR patches and form the super-resolved image $X$. 
We present this model in Fig.~\ref{fig:base}
and refer to it as Deep Multimodal Sparse Coding network (DMSC).

The proposed network can be trained end-to-end using 
the mean square error (MSE) loss function: 
\begin{equation}
\label{eq:obj}
\min_{\Theta} \sum\limits_{i} \| \hat{X}_i-X_i\|_2^2,
\end{equation}   
where, $\Theta$ denotes the set of all network parameters, 
$X_i$ is the corresponding ground-truth HR image of the target modality,
and $\hat{X}_i$ is the super-resolved estimation computed by the network.
\begin{figure}[t!]
	\centering
	\includegraphics[width=0.43\textwidth]{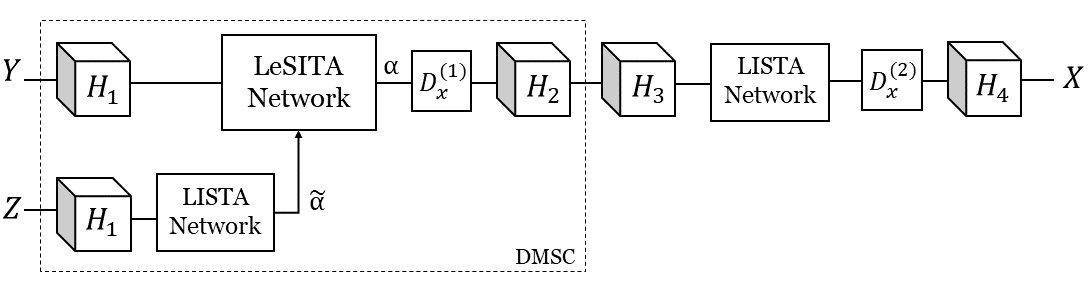}\\
	\caption{The proposed $\text{DMSC}_+$ model for multimodal image SR consists of the base DMSC model
	and a LISTA subnetwork with a linear decoder. 
	The convolutional layer $H_3$ performs patch extraction (similar to $H_1$), 
	while $H_4$ implements a patch aggregator layer.}
	\label{fig:deep}
\end{figure}
%--------------------------------------------------------------------%
\subsection{$\text{DMSC}_+$: Deep Multimodal Image SR Network}
%--------------------------------------------------------------------%
The DMSC network presented in Section~\ref{sec:base}
learns joint representations of three different image modalities,
that is, the input LR image $Y$, the guidance modality $Z$
and the HR image $X$.
Learning representations that encode only the common information 
among the different modalities is critical 
for the performance of the model.
Nevertheless, some information from the guidance modality may be misleading
when learning a mapping between the LR and HR versions of the target modality. 
In other words, the encoding performed by the LISTA branch may result in transferring
unrelated information to the LeSITA encoder.
As a result, the latent representation of the target modality may not capture the  
underlying mapping from the LR space to the HR space. 

As the performance of the model relies on the learned LR/HR transformation 
of the target modality, we present an architecture equipped with an uncoupling component
that focuses on learning the LR/HR transformation without using side information. 
The proposed framework consists of two different subnetworks:
(i) a DMSC subnetwork performing fusion of the information of the different image modalities,
and (ii) a subnetwork for the enhancement of the LR/HR transformation. 
The second is realized by a LISTA encoder followed by a linear decoder
and includes convolutional layers ($H_3$, $H_4$) to operate on the entire image.
The proposed deep multimodal framework, referred to as $\text{DMSC}_+$, is depicted in Fig.~\ref{fig:deep}.
The network is trained using objective~\eqref{eq:obj}.
\begin{table}[t!]
	\centering
	\caption{Performance comparison [in terms of PSNR (dB)] for $\times$2 SR\ upscaling (The results for scale $\times$2 are not presented in CDLSR \cite{MSR}). }
	\label{tab:table1}
	\begin{center}
%				\addtolength{\tabcolsep}{-1.7pt}
		\begin{tabular}{c || c | c | c | c |c}
			\hline
			
			$\times$2&{CSCN }&{ACSC }&{DJF }&\textbf{DMSC}&\textbf{$\text{DMSC}_+$}\\
			
			\hline
			\hline
			u-0006 & 39.47& 39.78 & 41.52& 41.79 &\textbf{43.21}\\
			u-0017 & 36.76& 36.64 & 38.65& 39.39 &\textbf{40.41}\\
			o-0018 & 33.98& 34.26 & 34.78& 36.02 &\textbf{37.90}\\
			u-0026 & 32.94& 33.11 & 33.15& 33.98 &\textbf{34.96}\\
			o-0030 & 33.34& 33.32 & 35.67& 36.32 &\textbf{37.73}\\
			u-0050 & 33.31& 33.39 & 32.60& 33.10 &\textbf{33.78}\\
			\hline
			\hline
			Average & 34.97& 35.09 & 36.07& 36.85&\textbf{37.99}\\
			\hline
			
		\end{tabular}
	\end{center}
\end{table}
\begin{table}[t!]
	\centering
	\caption{Performance comparison [in terms of PSNR (dB)] for $\times$4 SR\ upscaling. }
	\label{tab:table2}
	\begin{center}
%		\addtolength{\tabcolsep}{-1.7pt}
		\begin{tabular}{c || c | c | c | c | c |c}
			\hline
			
			$\times$4&{CSCN }&{ACSC }&{DJF }&{CDLSR}&\textbf{DMSC}&\textbf{$\text{DMSC}_+$}\\

			\hline
			\hline
			u-0006 & 32.60&  32.61& 36.04& 36.79& 37.24 &\textbf{37.82}\\
			u-0017 & 31.68&  31.66& 34.18& 35.27&  35.04 &\textbf{35.75}\\
			o-0018 & 27.28&  27.42& 30.72& \textbf{33.01}&  32.30&32.91\\
			u-0026 & 27.91&  27.92& 29.21& 30.35& 30.12&\textbf{30.40}\\
			o-0030 & 27.72&  27.66& 31.27& \textbf{32.71}&32.30  &32.66\\
			u-0050 & 28.20&  27.80& 28.58& 29.37&29.39 &\textbf{29.64}\\
			\hline
			\hline
			Average & 29.24&  29.18& 31.67& 32.92& 32.73 &\textbf{33.19}\\
			\hline
			
		\end{tabular}
	\end{center}
\end{table}
\begin{table}[t!]
	\centering
	\caption{Performance comparison [in terms of PSNR (dB)] for $\times$6 SR\ upscaling. }
	\label{tab:table3}
	\begin{center}
%				\addtolength{\tabcolsep}{-1.7pt}
		\begin{tabular}{c || c | c | c | c | c |c}
			\hline
			
			$\times$6&{CSCN }&{ACSC }&{DJF }&{CDLSR}&\textbf{DMSC}&\textbf{$\text{DMSC}_+$}\\
			
			\hline
			\hline
			u-0006 & 29.94& 29.97 & 34.92& 34.15&  35.43 &\textbf{35.74}\\
			u-0017 & 29.53& 29.48 & 32.80& 32.98&  33.09 &\textbf{33.55}\\
			o-0018 & 24.57& 24.70 & 29.92& 31.03& 30.44 &\textbf{31.34}\\
			u-0026 & 25.79& 25.97 & 28.38& 28.88&  28.77&\textbf{29.01}\\
			o-0030 & 25.86& 25.91 & 30.00& 30.52&  30.37&\textbf{30.61}\\
			u-0050 & 26.71& 26.43 & 27.64& 28.37&  28.27&\textbf{28.45}\\
			\hline
			\hline
			Average & 27.07& 27.08 & 30.62& 30.99&  31.06&\textbf{31.45}\\
			\hline
		\end{tabular}
	\end{center}
\end{table}
\vspace{-0.05cm}
%%%%%%%%%%%%%%%%%%%%%%%%%%%%%%%%%%
\section{Experiments}
\label{sec:experiment}
%%%%%%%%%%%%%%%%%%%%%%%%%%%%%%%%%%
In this section, we first report the implementation details of the proposed DMSC and $\text{DMSC}_+$ models. 
Then, we present experimental results on multimodal image SR.

%--------------------------------------------------------------------%
%\subsection{Implementation Details}
%--------------------------------------------------------------------%
The convolutional layers denoted by $H_1$ (patch extractors) 
are realized with $100$ filters of size $7\times7$
to extract $100$-dimensional feature vectors from patches of size $7\times7$
of the LR image and the HR side information. 
Each feature vector is then processed by the corresponding LeSITA or LISTA encoder 
to produce a sparse representation of the LR input and the side information. 
$H_3$ is realized in a similar way.
Each of the convolutional layers $H_2$ and $H_4$ contains one $5\times5\times49$ filter
to build-up the super-resolved NIR image from the computed patches. 
The sizes of the linear filters $W$ and $R$ are set to $128\times100$, while $S$ and $Q$ are set to $128\times128$. 
The linear decoder layer $D_x$ is realized by a $128\times49$ linear filter 
that recovers $7\times7$ HR patches. 
We note that all convolutional layers use padding 
such that the networks preserve the spatial size of the input. 
We use learnable scalars for the parameters $\gamma$ and $\mu$ of the proximal operators. 
We initialize the convolutional and linear filters using random weights drawn from 
a Gaussian distribution with standard deviation $0.1$. 
The parameters $\gamma$ and $\mu$ are initialized to $0.15$. 

%--------------------------------------------------------------------%
%\subsection{Experimental Results}
%--------------------------------------------------------------------%
We provide experimental results for the DMSC and $\text{DMSC}_+$ models,
and compare their performance with existing single-modal and multimodal SR methods.
In our experiments, we employ the EPFL RGB-NIR dataset\footnote{https://ivrl.epfl.ch/supplementary\_material/cvpr11/}. 
The dataset includes spatially aligned RGB and near-infrared (NIR) image pairs, 
capturing the same scene of $477$ different landscapes. 
Taking into account the high cost per pixel in NIR cameras, 
we want to super-resolve LR NIR images using the corresponding HR RGB image of the same scene as side information.
A preprocessing step involves upscaling the NIR LR image to the desired resolution using bicubic interpolation, 
which results in a blurry image given as input to the model.
We convert the RGB images to YCbCr and use only the luminance channel as side information. 
The training dataset contains $35000$ samples of NIR/RGB pairs; 
due to memory and computational limitations, we use image patches of size $60\times60$ to train our models.
We reserve $6$ image pairs for testing\footnote{A test image is identified by a letter \textquotedblleft{u}\textquotedblright, 
\textquotedblleft{o}\textquotedblright \ referring to the folders \texttt{urban} and \texttt{oldbuilding} in the dataset, 
and a code \textquotedblleft{00xx}\textquotedblright.}. 
 
\begin{figure*}
\centering
\subfloat[]{\includegraphics[scale=0.185]{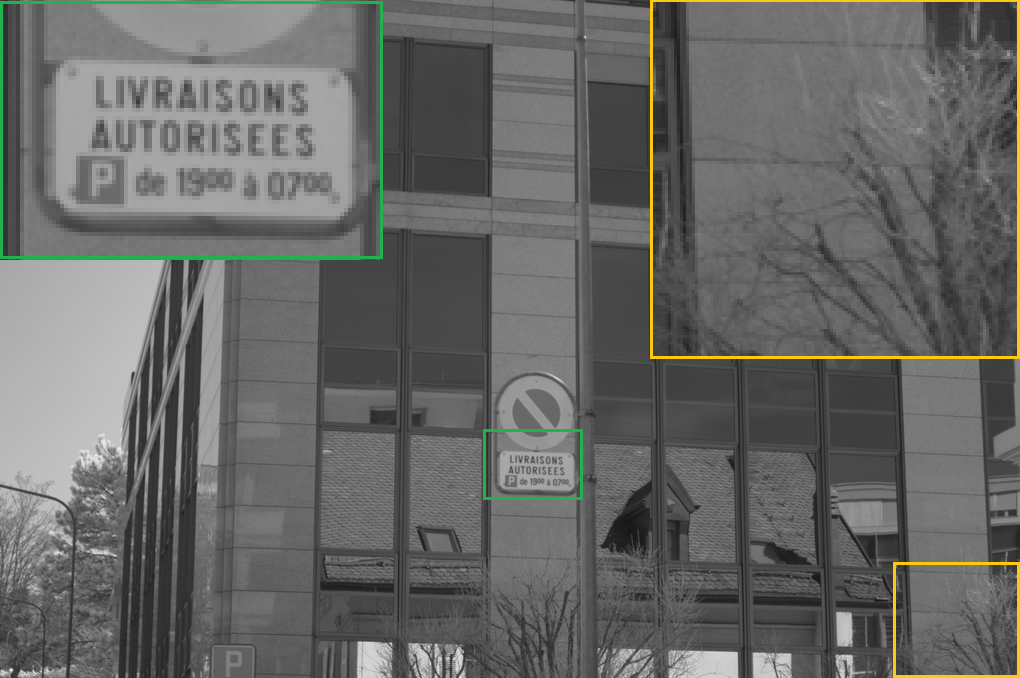}}
\hspace{0.05cm}
\subfloat[]{\includegraphics[scale=0.185]{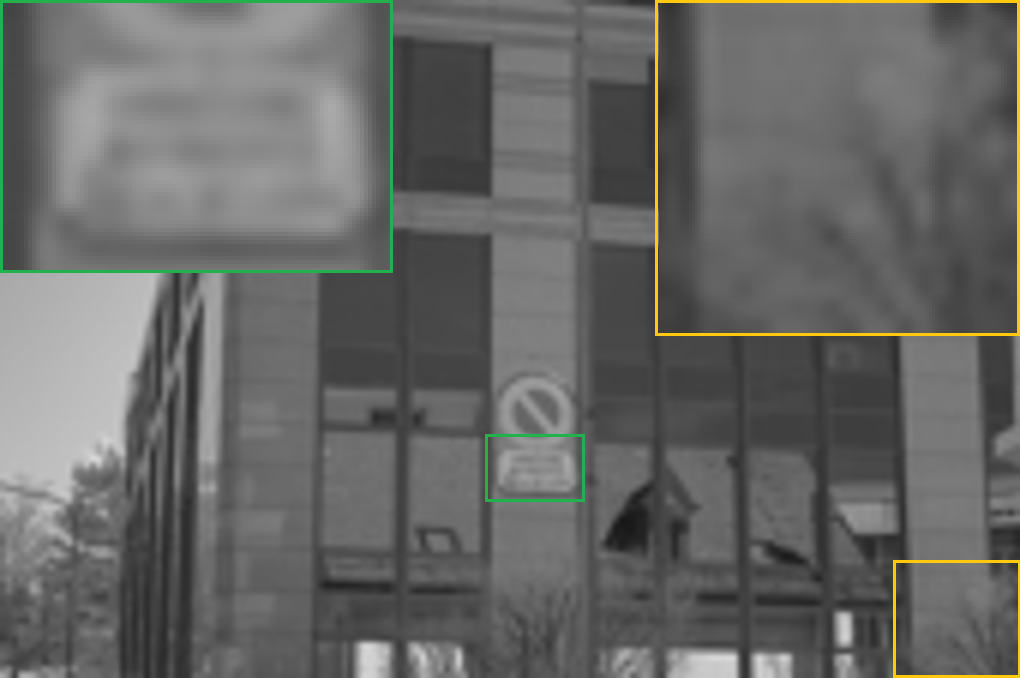}}
\hspace{0.05cm}
\subfloat[]{\includegraphics[scale=0.185]{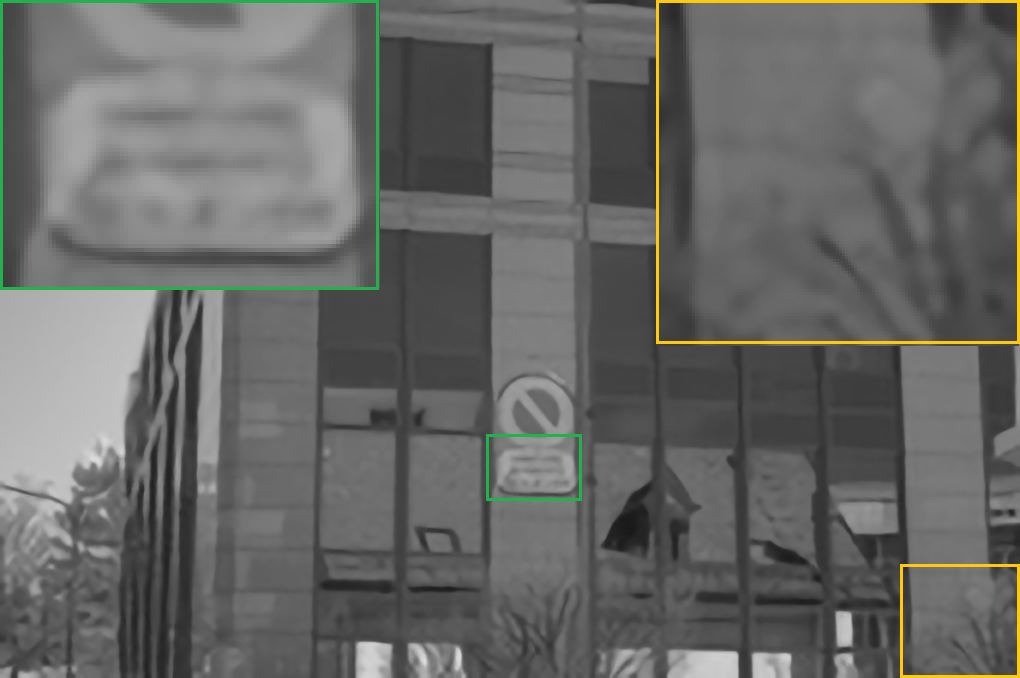}}
\hspace{0.05cm}
\subfloat[]{\includegraphics[scale=0.185]{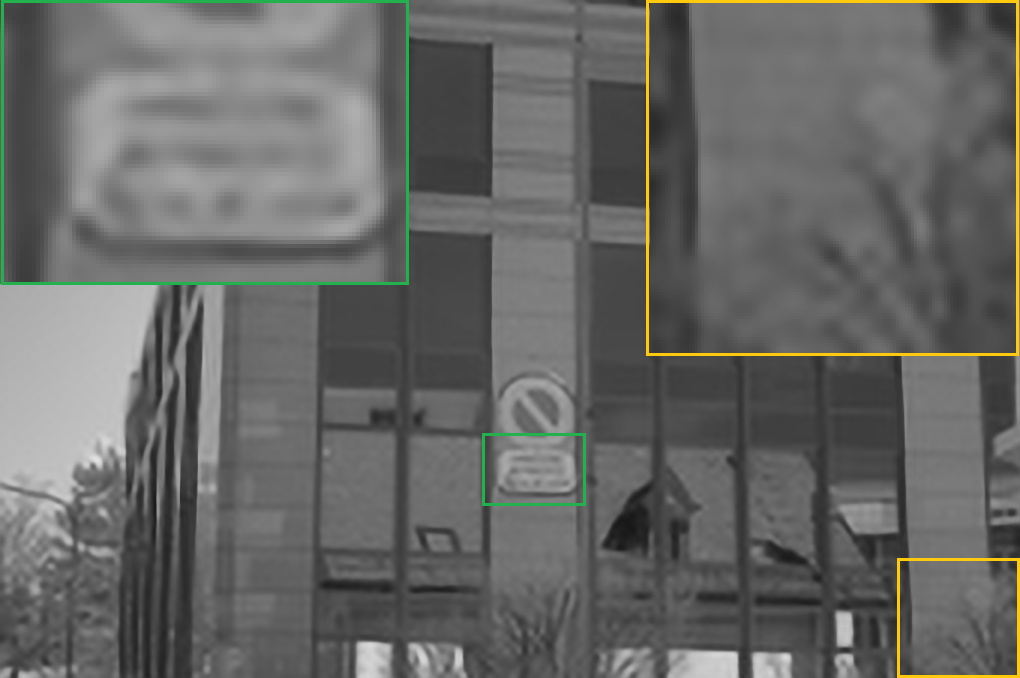}}
\hspace{0.05cm}
\subfloat[]{\includegraphics[scale=0.185]{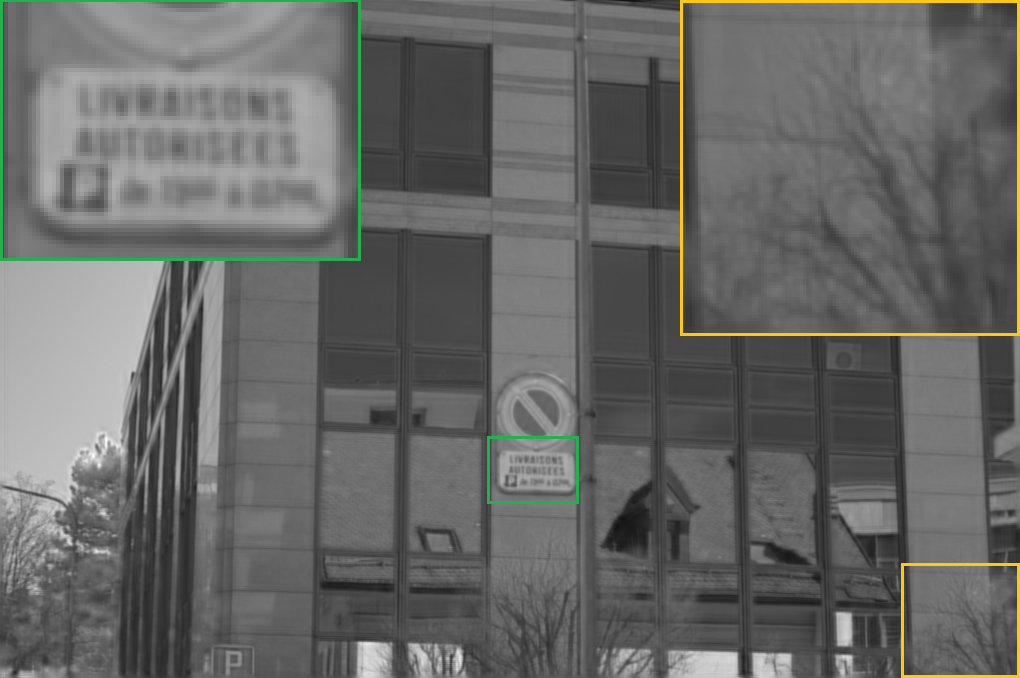}}
\hspace{0.05cm}
\subfloat[]{\includegraphics[scale=0.185]{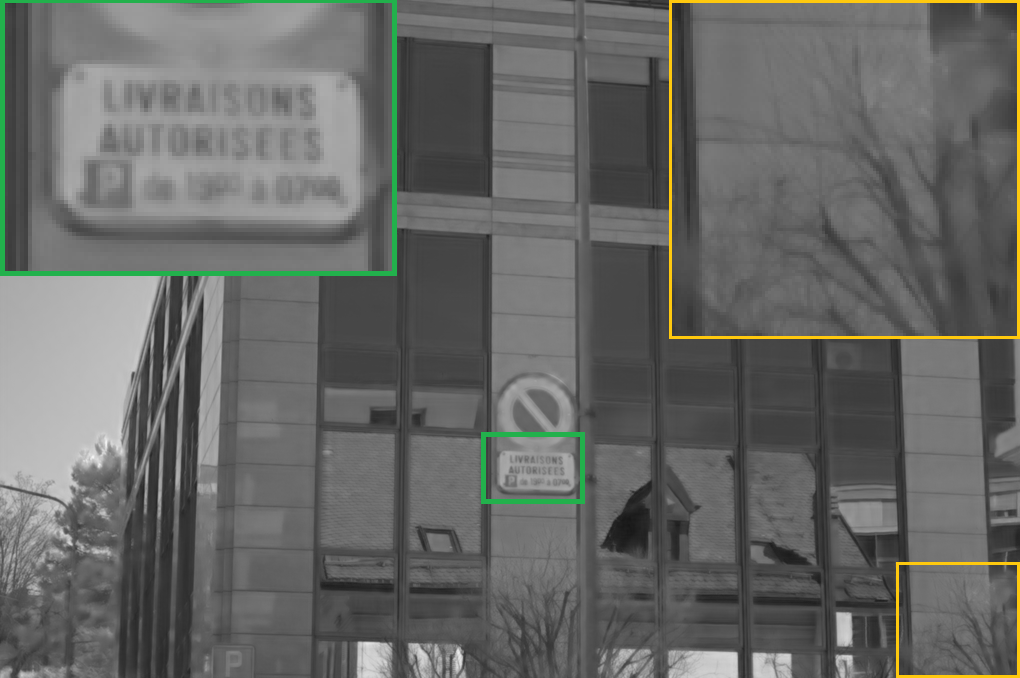}}

 \caption{$\times$6 upscaling for (a) the test image ``u0017" (ground-truth) with (b) bicubic, (c) CSCN~\cite{Huang}, (d) ACSC~\cite{raja}, (e) DJF~\cite{DJF} and (f) $\text{DMSC}_+$. Results for CDLSR~\cite{MSR} are not presented as the code is not available.}   
 \label{fig:visualEx}
\end{figure*}

We train the proposed models for three SR scales, $\times2$, $\times4$ and $\times6$,
by minimizing the objective~\eqref{eq:obj} utilizing ADAM optimizer. 
We compare the proposed models with 
(i) a coupled dictionary learning (CDLSR) method~\cite{MSR},
(ii) the deep joint image filtering (DJF) method~\cite{DJF}, 
(iii) an approximate convolutional sparse coding network (ACSC)~\cite{raja},
and (iv) a cascaded sparse coding network (CSCN)~\cite{Huang}. 
CDLSR and DJF perform multimodal image SR using sparse coding 
and convolutional neural networks, respectively. 
ACSC and CSCN are unimodal neural networks and do not use information from another image modality. 
Results in terms of Peak Signal-to-Noise Ratio (PSNR) for different scales 
are presented in Tables~\ref{tab:table1},~\ref{tab:table2} and~\ref{tab:table3}.
As can be seen, the DMSC model achieves state-of-the-art performance at most scaling factors. 
For instance, the average gains over the best competing method 
for $\times2$ and $\times6$ upscaling factors are 0.78 dB and 0.07 dB, respectively. 
However, for scale $\times4$ the average PSNR is 0.19 dB less than the best previous method. 
The $\text{DMSC}_+$ always outperforms existing techniques in terms of average PSNR 
and exhibits the best values for most of the testing images at all scales. 
A visual example presented in Fig.~\ref{fig:visualEx} corroborates our numerical results.
%

%%%%%%%%%%%%%%%%%%%%%%%%%%%%%%%%%%
\section{Conclusions}
\label{sec:conclusion}
%%%%%%%%%%%%%%%%%%%%%%%%%%%%%%%%%%
We developed two novel deep multimodal models, namely DMSC and $\text{DMSC}_+$, 
for the super-resolution of an LR image of a target modality 
with the aid of an HR image from another modality. 
The proposed design relies on the unfolding of an iterative algorithm 
for sparse approximation with side information.
The architecture incorporates sparse priors 
and effectively integrates the available side information. 
We applied the proposed models 
to super-resolve NIR\ images using RGB\ images as side information.
We compared our models with existing single-modal 
and multimodal designs, showing their superior performance.

%\vspace{-0.06cm}

\bibliographystyle{IEEEbib}
\bibliography{v4}

\begin{thebibliography}{10}

\bibitem{sparse1}
J.~Yang, J.~Wright, T.~S. Huang, and Y.~Ma,
\newblock ``Image super-resolution via sparse representation,''
\newblock {\em IEEE Transactions on Image Processing}, vol. 19, pp. 2861--2873,
  12 2010.

\bibitem{sparse2}
S.~Mallat and G.~Yu,
\newblock ``Super-resolution with sparse mixing estimators,''
\newblock {\em IEEE Transactions on Image Processing}, vol. 19, no. 11, pp.
  2889--2900, 2010.

\bibitem{sparse3}
J.~Yang, Z.~Wang, Z.~Lin, S.~Cohen, and T.~S. Huang,
\newblock ``Coupled dictionary training for image super-resolution,''
\newblock {\em IEEE Transactions on Image Processing}, vol. 21, no. 8, pp.
  3467--3478, 2012.

\bibitem{10prim}
W.~Han, S.~Chang, D.~Liu, M.~Yu, M.~Witbrock, and T.~S. Huang,
\newblock ``Image super-resolution via dual-state recurrent networks,''
\newblock in {\em IEEE Conference on Computer Vision and Pattern Recognition
  (CVPR)}, 2018.

\bibitem{11prim}
M.~Haris, G.~Shakhnarovich, and N.~Ukita,
\newblock ``Deep back-projection networks for super-resolution,''
\newblock in {\em IEEE Conference on Computer Vision and Pattern Recognition
  (CVPR)}, 2018.

\bibitem{15prim}
J.~Kim, J.~K. Lee, and K.~M. Lee,
\newblock ``Accurate image super-resolution using very deep convolutional
  networks,''
\newblock in {\em IEEE Conference on Computer Vision and Pattern Recognition
  (CVPR)}, 2016.

\bibitem{17prim}
W.~S. Lai, J.~B. Huang, N.~Ahuja, and M.~H. Yang,
\newblock ``Deep laplacian pyramid networks for fast and accurate
  super-resolution,''
\newblock in {\em IEEE Conference on Computer Vision and Pattern Recognition
  (CVPR)}, 2017.

\bibitem{19prim}
B.~Lim, S.~Son, H.~Kim, S.~Nah, and K.~M. Lee,
\newblock ``Enhanced deep residual networks for single image
  super-resolution,''
\newblock in {\em IEEE Conference on Computer Vision and Pattern Recognition
  Workshop (CVPRW)}, 2017.

\bibitem{Huang}
D.~Liu, Z.~Wang, B.~Wen, J.~Yang, W.~Han, and T.~S. Huang,
\newblock ``Robust single image super-resolution via deep networks with sparse
  prior,''
\newblock {\em IEEE Transactions on Image Processing}, vol. 25, no. 7, pp.
  3194--3207, 2016.

\bibitem{LISTA}
K.~Gregor and Y.~LeCun,
\newblock ``Learning fast approximations of sparse coding,''
\newblock in {\em IEEE International Conference on Machine Learning (ICML)},
  2010.

\bibitem{MSR2}
K.~He, J.~Sun, and X.~Tang,
\newblock ``Guided image filtering,''
\newblock {\em IEEE Transactions on Pattern Analysis and Machine Intelligence},
  vol. 35, no. 6, pp. 1397--1409, 2013.

\bibitem{DJF}
Y.~Li, J.~B. Huang, N.~Ahuja, and M.~H. Yang,
\newblock ``Deep joint image filtering,''
\newblock in {\em European Conference on Computer Vision (ECCV)}, 2016, vol.
  9908.

\bibitem{MSR}
P.~Song, X.~Deng, J.~F.~C. Mota, N.~Deligiannis, P.~L. Dragotti, and M.~R.~D.
  Rodrigues,
\newblock ``Multimodal image super-resolution via joint sparse representations
  induced by coupled dictionaries,''
\newblock {\em IEEE Transactions on Computational Imaging}, 2019.

\bibitem{MSR1}
J.~Kopf, M.~F. Cohen, D.~Lischinski, and M.~Uyttendaele,
\newblock ``Joint bilateral upsampling,''
\newblock {\em ACM Transactions on Graphics}, vol. 26, no. 3, 2007.

\bibitem{MSR3}
B.~Ham, M.~Cho, and J.~Ponce,
\newblock ``Robust guided image filtering using nonconvex potentials,''
\newblock {\em IEEE Transactions on Pattern Analysis and Machine Intelligence},
  vol. 40, no. 1, pp. 192--207, 2018.

\bibitem{MSR4}
S.~Gu, W.~Zuo, S.~Guo, Y.~Chen, C.~Chen, and L.~Zhang,
\newblock ``Learning dynamic guidance for depth image enhancement,''
\newblock in {\em IEEE Conference on Computer Vision and Pattern Recognition
  (CVPR)}, 2017.

\bibitem{EVA}
E.~Tsiligianni and N.~Deligiannis,
\newblock ``Learning fast sparse representations with the aid of side
  information,''
\newblock Tech. {R}ep., 2019,
\newblock [Online Available: {https://bit.ly/2WIJB5z}].

\bibitem{tropp2010computational}
J.A. Tropp and S.J. Wright,
\newblock ``Computational methods for sparse solution of linear inverse
  problems,''
\newblock {\em Proceedings of the IEEE}, vol. 98, no. 6, pp. 948--958, 2010.

\bibitem{ISTA}
I.~Daubechies, M.~Defrise, and C.~De Mol,
\newblock ``An iterative thresholding algorithm for linear inverse problems
  with a sparsity constraint,''
\newblock {\em Communications on Pure and Applied Mathematics}, vol. 57, 11
  2004.

\bibitem{xin2016maximal}
B.~Xin, Y.~Wang, W.~Gao, D.~Wipf, and B.~Wang,
\newblock ``Maximal sparsity with deep networks?,''
\newblock in {\em Advances in Neural Information Processing Systems (NIPS)},
  2016, pp. 4340--4348.

\bibitem{borgerding2017amp}
M.~Borgerding, P.~Schniter, and S.~Rangan,
\newblock ``{AMP}-inspired deep networks for sparse linear inverse problems,''
\newblock {\em IEEE Transactions on Signal Processing}, vol. 65, no. 16, pp.
  4293--4308, 2017.

\bibitem{prox}
C.~Bertocchi, E.~Chouzenoux, M.-C. Corbineau, J.-C.Pesquet, and M.~Prato,
\newblock ``Deep unfolding of a proximal interior point method for image
  restoration,''
\newblock {\em CoRR}, vol. abs/1812.04276, 2018.

\bibitem{NikosIT}
J.~F.~C. Mota, N.~Deligiannis, and M.~R.~D. Rodrigues,
\newblock ``{Compressed Sensing with Prior Information: Strategies, Geometry,
  and Bounds},''
\newblock {\em IEEE Transaction on Information Theory}, vol. 63, pp.
  4472--4496, 2017.

\bibitem{nikosGlobsip}
J.~F.~C. {Mota}, N.~{Deligiannis}, and M.~R.~D. {Rodrigues},
\newblock ``Compressed sensing with side information: Geometrical
  interpretation and performance bounds,''
\newblock in {\em GlobalSIP}, 2014, pp. 512--516.

\bibitem{Bach2012}
F.~Bach, R.~Jenatton, J.~Mairal, and G.~Obozinski,
\newblock ``{Optimization with Sparsity-Inducing Penalties},''
\newblock {\em Foundations and Trends in Machine Learning}, vol. 4, no. 1, pp.
  1--106, Jan. 2012.

\bibitem{raja}
H.~Sreter and R.~Giryes,
\newblock ``Learned convolutional sparse coding,''
\newblock in {\em IEEE International Conference on Acoustics, Speech, and
  Signal Processing (ICASSP)}, 2018, pp. 2191--2195.

\end{thebibliography}

\end{document}